\documentclass{article} 
\usepackage{iclr2020_conference,times}

\usepackage{amsmath,amsfonts,bm}

\def\eqref#1{equation~\ref{#1}}

\def\1{\bm{1}}

\def\vx{{\bm{x}}}

\DeclareMathAlphabet{\mathsfit}{\encodingdefault}{\sfdefault}{m}{sl}
\SetMathAlphabet{\mathsfit}{bold}{\encodingdefault}{\sfdefault}{bx}{n}

\def\sF{{\mathbb{F}}}

\usepackage{hyperref}
\usepackage{url}
\usepackage{xcolor}
\usepackage{graphicx}
\usepackage{multirow}

\title{Bayesian aggregation improves traditional single image crop classification approaches}

\author{Ivan Matvienko, Mikhail Gasanov, Anna Petrovskaia, \\ \bf Raghavendra Belur Jana, Maria Pukalchik, Ivan Oseledets 
\\
Center for Computational and Data-Intensive Science and Engineering (CDISE)\\
Skolkovo Institute of Science and Technology (Skoltech)\\
Moscow, Russia \\
\texttt{\{ivan.matvienko,mikhail.gasanov,anna.petrovskaia\}@skoltech.ru} \\
\texttt{\{r.jana,m.pukalchik,i.oseledets\}@skoltech.ru}
}
\iclrfinalcopy 
\begin{document}

\maketitle

\begin{abstract}
Machine learning (ML) methods and neural networks (NN) are widely implemented for crop types recognition and classification based on satellite images. However, most of these studies use several multi-temporal images which could be inapplicable for cloudy regions. We present a comparison between the classical ML approaches and U-Net NN for classifying crops with a single satellite image. The results show the advantages of using field-wise classification over pixel-wise approach. We first used a Bayesian aggregation for field-wise classification and improved on 1.5\% results between majority voting aggregation. The best result for single satellite image crop classification is achieved for gradient boosting with an overall accuracy of 77.4\% and macro F1-score 0.66. 
\end{abstract}

\section{Introduction}
Crop identification from satellite images is an essential task of precision agriculture. The accurate information about the growing crops provides the possibility to regulate agricultural products’ internal stocks and to draw strategies for the negotiation of agricultural commodities in financial markets. Recent developments in remote sensing approaches and data processing techniques have enabled both researchers and practitioners to simplify the process of crop identification.

The growing use of satellite imagery at very high spatial and temporal resolution enables land managers to obtain a variety of information on how land is used. A variety of methods for detecting crop patterns by classifying multi-source and multi-temporal data have been proposed and evaluated by using time-series from different satellites in recent years~\citep{ma2017review}. However, majority if this studies revealed some drawbacks, mainly attributed to the high cost of data collection for time-series, and cloud cover problem, which commonly observed in different regions: e.g., in northern countries a snow cover may occur more than four months per year and even during the vegetation season the total amount of sunny days may be less than 30\%; and in equatorial countries cloud cover may lead to 45 - 50\% during the year. That means that sometimes we can receive and analyze only one or two cloud-free images during the vegetation period from satellites. Thus, a sufficient single image crop classification approach need to be conducted for this purposes.

In this work, we apply different classical machine learning (ML) methods and Deep learning (DL) algorithm like convolutions neural networks (CNNs) to map crop type from a single Sentinel-2 satellite image in field-scale before and after different pixel aggregation. This work also contributes a novel approach to aggregate results of a pixel-wise classification towards field-wise ones by using the Bayesian strategy. To the best of our knowledge, we give the first experimental application of aggregation strategy to ML and DL algorithms, which allow improving the overall classification performance significantly for crop type mapping of smallholder farms in Africa, even based on a single satellite image.

\section{Related works}

Several major approaches are used to tackle the problem of crop classification based on satellite images. For the last decade, the most popular and powerful instrument for this case were classical machine learning (ML) techniques. Many authors have already investigated crop-type classification using traditional statistical or machine learning methods: random forest \citep{schultz2015self, vuolo2018much, shukla2018performance}, discriminant analysis \citep{arafat2013crop}, k-nearest neighbor, extreme learning machine \citep{sonobe2017assessing}, Maximum Likelihood Classification \citep{lussem2016combined, arvor2013mapping}, CART decision trees \citep{shukla2018performance, chen2018mapping} like support vector machine (SVM) \citep{zheng2015svm, kang2018svm, gilbertson2017effect, asgarian2016crop, kumar2017statistical, lussem2016combined}. The best results of crop recognition was shown with a help of  SVM where researchers achieved a 75\% overall accuracy (OA) \citep{khatami2016meta}.  
Recently, much attention is paid to the deep learning (DL) methods; e.g., a deep convolutional neural network \citep{kussul2017deep, ji20183d, zhong2019deep}, and LSTM recurrent neural networks \citep{zhong2019deep} were successfully applied for crop recognition. It is possible to achieve similar results as SVM with approximately the same efficiency - 74\% of OA, according to \citet{khatami2016meta}. However, all of the mentioned above studies were focused on time-series analyzing, so they efficiently and OA to single-image crop recognition is lower or not checked. 

\section{The "Farmpins Crop Classification" dataset and Features}

We used the dataset for South Africa from 'Farmpins Crop Classification' Challenge\footnote{\url{https://zindi.africa/competitions/farm-pin-crop-detection-challenge}}. The dataset consists of 2497 label fields for nine crop classes and Sentinel 2B satellite images. The present work focused on a mono-temporal approach (acquisition dates centered on the 01 January 2017 when the majority of crops were near peak greenness). We upsampled all satellite images bands to 10 m spatial resolution using bilinear interpolation.
As input features, we used all 13 bands from satellite images and additionally generated the following vegetation indices: Normalized Difference Vegetation Index (NDVI), Enhanced Vegetation Index (EVI), Normalized Difference Red Edge (NDRE) and Modified Soil-adjusted Vegetation Index (MSAVI). We also normalized features to zero mean and unit variance. For training and validation datasets we split fields into train, validation, and test parts as 75\%, 12.5\%, and 12.5\% of the whole number of fields, such a way, to preserve the proportions of pixels of each crop equal between splits.

\section{Experiments}
 

\subsection{Classification methods}

For crop classification, we used the following methods: K-Nearest-Neighbors (KNN), Random Forest classifier (RF), Gradient Boosting classifier (GB). Even though we concentrated on classical ML algorithms, we also applied artificial neural network with U-Net architecture and SE-blocks ~\citep{ronneberger2015u}. We maximized the robustness of training algorithms by using different well-known approaches as Random Over- and Under-Sampling (ROS and RUS), Synthetic Minority Over-sampling Technique (SMOTE) \citep{chawla2002smote}, and simple weighting of classification classes.  The architecture of the U-Net and SE-blocks network is shown in the figure below. In order to evaluate the performance of the ML pipeline, we focused on improving such metrics as overall accuracy (OA) and macro F1-score.

\begin{figure}[!h]
\begin{center}
\includegraphics[width=4.5in]{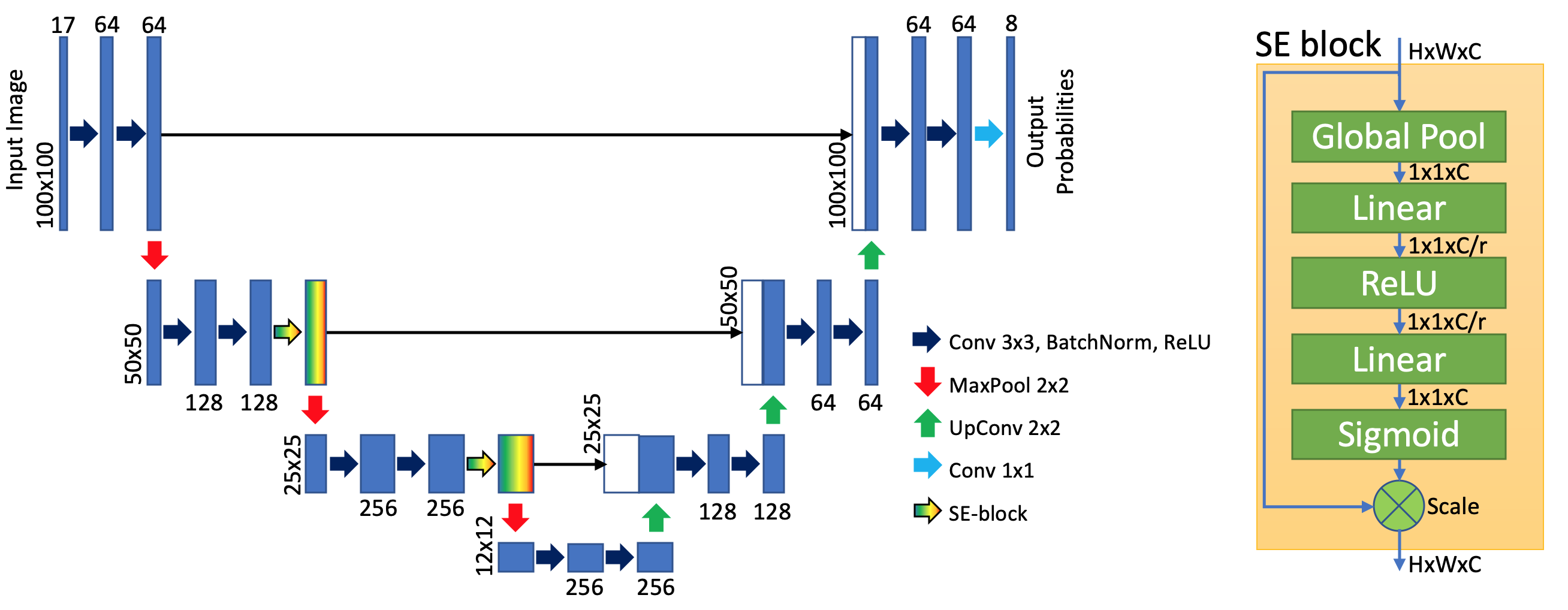}
\end{center}
\caption{Architecture of UNet with SE-blocks}
\end{figure}

\subsection{Pixel aggregation methods}

We used the pixel-wise approach output with the probabilities $p(k|\vx)$ of a particular pixel was represented by feature vector $\vx$ to the class $k$. Meanwhile, the results of the field classification task provide more preferable data for farmers and decision-makers than results of pixel-based one; that’s why we focused on this type of crop classification. As the classification method has some misclassification rates, some number of pixels within the fields were assigned to the wrong class. Still, there is prior information that all pixels from a particular field belong to the same class. The natural way to get rid of miss classified pixels is to assign all pixels within one field to one class. To perform it we apply three methods: majority voting, average voting, and Bayesian aggregation.

The procedure of majority aggregation is quite simple: for each pixel it chooses the class $c$ with the highest predicted probability  $c = \arg\max_k p(k|\vx)$. Then the whole is assigned to the crop with the highest number of pixels predicted. For the averaging voting, we are firstly averaging the probabilities for each class independently within the field and then picking the class with the highest mean probability. However, the majority voting approach might be inadequate in the case of small fields represented by a few numbers of pixels. The problem with the aggregating strategies described above is that it combines the probabilities in a wrong manner: 2 consequent positive predictions should increase the confidence - the overall probability should be greater of maximal ones and vice versa. 


The natural way of incorporating the prior information that all pixels in the same field have the same class is Bayes theorem. Let $\sF$ be a set of features of pixels from some field: $\vx_i \in \sF$. Summing up the log-odds of the predictions from every pixel (\ref{log-odds}), final decision is obtained using (\ref{final-des}):
\begin{equation}
\label{log-odds}
    I(k) = \sum_{\vx \in \sF} \log \frac{1 - p(k|\vx_i)}{p(k|\vx_i)}
\end{equation}
\begin{equation}
\label{final-des}
    c = \arg \max_k \frac{1}{1 + \exp({I(k)})}
\end{equation}
We found that before applying the Bayesian aggregation, it is useful to perform smoothing of raw predictions in the following way:
\begin{equation}
    \widehat{p}_{_\alpha}(k|\vx_i) = \alpha \, p(k|\vx_i) + \frac{1 - \alpha}{N - 1} (1 - p(k|\vx_i))
\end{equation}
where $\alpha \in (0, 1)$ is a smoothing factor, $N$ is the number of classes for classification. The choice of factor $\alpha$ depends on the methods used. We conducted a little grid search and found that for $N=8$ as in our case values $ \alpha \in (0.3, 0.4)$ perform best. 


\section{Results and Discussion}

Procedures explained in Section 3 have been implemented on single-image crop-recognition. Table \ref{p-table} shows the main results of our pixel-wise prediction before the aggregation procedure for all tested algorithms.

\begin{table}[ht]
\caption{Comparison of pixel-wise classification results for tested approaches before aggregation}
\label{p-table}
\begin{center}
\begin{tabular}{lcccccccc} \multirow{2}{*}{\bf Classifier}                             & \multicolumn{2}{c}{\bf ROS} &                                 \multicolumn{2}{c}{\bf RUS} &                                 \multicolumn{2}{c}{\bf SMOTE} &                               \multicolumn{2}{c}{\bf Weighting}\\
 & OA   & Macro F1 & OA & Macro F1 & OA & Macro F1 & OA & Macro F1 \\ \hline
KNN & 53.0  & 0.39  & 43.5  & 0.33  & 53.3  & 0.40  & ---   & ---   \\
RF  & 62.7  & 0.46  & 50.3  & 0.37  & 62.4  & 0.46  & 62.5  & 0.46  \\
GB  & \bf 68.5  & \bf 0.51  & 55.3  & 0.42  & \bf 65.7  & \bf 0.50  &     \bf 68.7  & \bf 0.50   \\
U-Net+SE  & ---  & ---  & ---  & ---  & ---  & ---  & \bf 70.1   & \bf 0.57  
\end{tabular}
\end{center}
\end{table}

We see that different models have very different OA for crop classification; e.g., U-Net neural network with SE-blocks (U-Net+SE) outperformed other methods like GB, KNN classifier, and RF. Meanwhile, all tested models cannot outperform the previous results, achieved by \citet{m2019semantic}.  Table \ref{agg-table} presents a comparison between different aggregation techniques performance to field-based crop classification. 
\begin{table}[h!]
\caption{Comparison of different aggregation techniques for field-based crop classification by ML and DL methods}
\begin{center}
\label{agg-table}
\begin{tabular}{llcccc}
\multicolumn{2}{c}{}&                           \multicolumn{4}{c}{\bf Overall Accuracy (\%)}\\
{} & Aggregation & {KNN} & {RF} & {GB} & {U-Net+SE} \\ \hline 
\multirow{3}{*}{\begin{tabular}[c]{@{}l@{}}\bf ROS \end{tabular}} 
 & Bayesian  & \bf 72.66 & \bf 73.61 & \bf 77.44 & ---  \\
 & Averaging & 72.28 & 73.42 & 77.25 & ---  \\
 & Majority  & 72.08 & 72.08 & 76.67 & ---  \\ \hline
\multirow{3}{*}{\begin{tabular}[c]{@{}l@{}}\bf RUS \end{tabular}} 
 & Bayesian  & 61.41 & \bf 60.11 & \bf 70.44 & ---  \\
 & Averaging & 61.41 & 59.92 & 70.08 & ---  \\
 & Majority  & 58.62 & 58.48 & 69.69 & ---  \\ \hline
\multirow{3}{*}{\bf SMOTE}                     
 & Bayesian  & 72.85 & \bf 73.23 & \bf77.44 & ---  \\
 & Averaging & 73.04 & 73.04 & 77.44 & ---  \\
 & Majority  & 72.66 & 71.70 & 75.72 & ---  \\ \hline
\multirow{3}{*}{\bf Weighting}                            & Bayesian  & ---  & \bf 72.85 & \bf 77.44 & \bf 71.58 \\
 & Majority  & ---  & 72.47 & 77.06 & 71.34 \\
 & Averaging & ---  & 71.13 & 75.91 & 71.22
\end{tabular}
\end{center}
\end{table}
The effect of aggregation on the results is promising: we have applied three different aggregation approaches, and all of them showed marked improvement of crop recognition results for field scale, and the Bayesian aggregation showed the best OA between tested methods. Interestingly, the U-Net application provides similar OA results for both pixel-wise and field-wise crop classification, probably because of U-net consider particular contextual information. 
The results of field-based classification after the Bayesian aggregation approach for tested methods are presented in Table \ref{f-table}. According to our results RF, KNN, and GB were sufficient for crop classification for a single satellite image. It also appears that ROS and weighting techniques produce the best highest overall accuracy and macro F1 score between tested methods.  To the best of our knowledge, this is a first study where the Bayesian aggregation approach was successfully applied to improve crop classification in field scale. 
\begin{table}[h!]
\caption{Bayesian aggregation improved performance of ML approaches for field-wise classification task}
\label{f-table}
\begin{center}
\begin{tabular}{lcccccccc}
           \multirow{2}{*}{\bf Classifier }& \multicolumn{2}{c}{\bf ROS} & \multicolumn{2}{c}{\bf RUS} & \multicolumn{2}{c}{\bf SMOTE} & \multicolumn{2}{c}{\bf Weighting} \\
& OA & Macro F1 & OA & Macro F1 & OA & Macro F1 & OA & Macro F1  \\ \hline 
KNN & 72.2 & 0.59 & 61.4 & 0.49 & 72.8 & 0.61 & --- & ---  \\
RF  & 73.0 & 0.61 & 59.4 & 0.47 & 73.2 & 0.47 & 72.7 & 0.61 \\
GB  & \bf 77.2 & \bf 0.66 & 70.3 & 0.58 & \bf 77.4 & \bf 0.66 & \bf 77.1 & \bf 0.63 \\
U-Net+SE  & ---  & ---  & ---  & --- & --- & --- & 71.5 & 0.58           
\end{tabular}
\end{center}
\end{table}

\section{Conclusion}
We compared the performance of classical machine learning methods for crop classification task in South Africa and were able to achieve reasonable performance with the use of aggregation strategy. We evaluated and compared the performance of different classical aggregation strategies and suggested a new one based on the Bayes theorem, which has never been used. It outperformed other aggregation strategies as majority voting for 1.5\% and averaging approach for 0.6\%. We believe, our achievements will help governments of developing countries for better agriculture regulation. We hope that the single image crop classification approach might be useful for the sustainable development of smallholder farmers.


\bibliography{iclr2020_conference}
\bibliographystyle{iclr2020_conference}

\appendix
\section{Appendix}
\subsection{Technical details}
The dataset from 'Farmpins Crop Classification' Challenge consists of 2497 labels fields of 9 classes presented in Table ~\ref{class-balance}.
\begin{table}[ht]
\caption{Number of fields of each crop after preprocessing}
\label{class-balance}
\begin{center}
\begin{tabular}{lllllllll}
\bf Cotton & \bf Dates & \bf Grass &  \bf Lucern & \bf Maize & \bf Pecan & \bf Vacant & \bf Vineyard & \bf Intercrop \\
\hline \\
113 & 2 & 85 & 468 & 251 & 135 & 233 & 789 & 71
\end{tabular}
\end{center}
\end{table}

For training, we have filtered out the fields with the area less than 0.5 ha (50 pixels). We also removed Dates class from the training set because there were only two fields of that crop in the dataset, and it is quite challenging to train the classifier with that amount of data. We also excluded 'Intercrop' class from the training set because this class is a combination of two other crops: Vineyard and Pecan, which are present in our dataset. After all these preprocessing steps, there are 2147 fields left.  

We used Sentinel 2 satellite images. Each image contains 13 channels (bands) of positive integer two-dimensional arrays. Bands have 3 different spatial resolutions: 60 m, 20 m, and 10 m.

For KNN, we used eight neighbors, and for each neighbor, we assigned an importance weight, which is inversely proportional to the distance between objects. For RF, we used an ensemble of 200 decision trees with a maximum depth of 15 and at least 10 samples in the leaf. We also boosted a classifier of 250 decision trees of 10 maximum depth. We conducted a boosting procedure with the sampling technique of 0.5 rate (used 50\% of dataset for every boosting iteration) and a learning rate of 0.05. U-Net network was trained using Adam optimization algorithm with an exponentially decaying strategy for learning rate with an initial one of $0.01$ and $0.977$ decrease rate for every epoch, and weight decay of $1e^{-4}$. For the U-Net application, we split each initial image with generated vegetation indices into patches of 100x100. For data augmentation, random rotation, shifting, and horizontal flipping were used with preliminary mirror padding.

\end{document}